\documentclass[10pt, a4paper]{article}

\usepackage[final]{lrec2026} 
\usepackage{amsmath}
\usepackage{booktabs}
\usepackage{multirow}
\usepackage{graphicx}
\usepackage{hyperref}

\title{EnTaCs: Analyzing the Relationship Between Sentiment and Language Choice in English-Tamil Code-Switching}

\name{Paul Bontempo}

\address{University of Colorado, Boulder \\
         Depts. of Computer Science and Linguistics \\
         paul.bontempo@colorado.edu}

\abstract{
This paper investigates the relationship between utterance sentiment and language choice in English-Tamil code-switched text, using methods from machine learning and statistical modelling. 
We apply a fine-tuned XLM-RoBERTa model for token-level language identification on 35,650 romanized YouTube comments from the DravidianCodeMix dataset, producing per-utterance measurements of English proportion and language switch frequency. 
Linear regression analysis reveals that positive utterances exhibit significantly greater English proportion (34.3\%) than negative utterances (24.8\%), and mixed-sentiment utterances show the highest language switch frequency when controlling for utterance length. 
These findings support the hypothesis that emotional content demonstrably influences language choice in multilingual code-switching settings, due to socio-linguistic associations of prestige and identity with embedded and matrix languages.
\\ \newline  }

\begin{document}

\maketitleabstract

\section{Introduction}

Code-switching is the phenomenon wherein bilingual speakers alternate between languages, dialects, or registers within a conversation or utterance \citep{poplack1980sometimes}. 
Research distinguishes two varieties by granularity of alternation. 
\textit{Inter-discursive} code-switching occurs only at utterance or discourse boundaries, often in response to changes in interlocutor, setting, or topic. 
\textit{Intra-discursive} code-switching—also termed code-mixing—involves alternation at sub-utterance boundaries (sentence, word, morpheme, or character level) and is typically an unconscious process reflecting speaker-internal attitudes rather than external pressures \citep{muysken2000bilingual}.

In code-switching, the normative or privileged language is the \textit{matrix language}, while the marked or minority language is the \textit{embedded language} \cite{MyersScotton2002}. 
The acceptability of utterances in a code-switching context is bound by morphosyntactic and phonotactic constraints as rigorously as in any monolingual utterance, though the constraints themselves are not simply a combination of the constituent languages' grammars \cite{Sankoff1981}; 
code-switching is a form of linguistic expression as complex and valid as any single language (although it is often stigmatized), and code-switching speakers follow internally consistent rules of grammaticality which govern where and how language switches can occur in an utterance.

In southern Indian states such as Tamil Nadu, English functions as a lingua franca and carries associations of academic and professional prestige, while Tamil carries emotional and identity-marking associations for native speakers \citep{eldho2023english}. 
Bilingualism is widespread in India, with approximately 12\% of the population reporting L2 English proficiency alongside a native Indian language \citep{sarma2025language}.

This sociolinguistic asymmetry motivates our line of inquiry: if bilingual speakers associate English with prestige and positive affect while associating Tamil with emotional weight and intimate expression, does the sentiment of an utterance predict the speaker's language choice when producing that utterance?
From this premise, our two specific research questions follow naturally:

\noindent\textbf{RQ1:} Does overall utterance sentiment predict the proportion of English (embedded language) relative to Tamil (matrix language) in code-switched utterances?

\noindent\textbf{RQ2:} Does overall utterance sentiment predict the frequency of language switches (alternation between English and Tamil) within an utterance?

\section{Related Work}

\paragraph{Code-mixed NLP for South Asian languages.}
The \texttt{DravidianCodeMix} dataset \citep{chakravarthi2022dravidian} provides manually annotated YouTube comments in Tamil-English, Kannada-English, and Malayalam-English, with automatically-applied but manually-verified labels for both sentiment and offensive language. 
It constitutes the largest publicly available resource for Dravidian code-mixed NLP and forms the basis for our analysis. Subsequent shared-task work on this corpus has focused primarily on \textit{classifying} sentiment in code-mixed text; 
\citet{babu2021sentiment} achieved state-of-the-art performance on Tamil-English sentiment classification in a shared task using a paraphrase-tuned XLM-RoBERTa model. 
Our work departs from this line by instead treating sentiment as a \textit{predictor} of linguistic behavior rather than as a target label, shifting the research question from classification accuracy towards sociolinguistic analysis. 
Broader surveys of NLP methods for code-switching \citep{sitaram2019survey} point towards token-level language identification as a critical enabling task, since most downstream analysis requires knowing which language each token belongs to.

\paragraph{Language identification in code-switched text.}
Token-level language identification in code-switched text is a challenging task unto itself, because romanized Tamil text shares script with English, preventing traditional language models from discriminating between the languages at the character level. 
Multilingual pre-trained models such as XLM-RoBERTa \citep{conneau2020unsupervised} have shown strong performance on cross-lingual token classification, making them well-suited for fine-tuning on this task. 
The LinCE benchmark \citep{aguilar2020lince} provides standardized evaluation for code-switching tasks including language identification.

\paragraph{Code-switching and sentiment.}
Prior work has noted that code-switching can be used metalinguistically to signal speaker attitude, group identity, and emotional stance \citep{yoder2017code}. 
The discursive asymmetry between prestige and heritage languages, with colonial languages signaling authority as evidence of wealth and foreign education while native languages carry emotional or personal influence, provides a sociolinguistic basis for our hypothesis that sentiment predicts language choice.

\section{Dataset}

\subsection{Source and Preprocessing}

We use the Tamil-English split of the DravidianCodeMix corpus \citep{chakravarthi2022dravidian}, which contains $44,161$ YouTube comments annotated with five utterance-level sentiment categories: \textit{Positive}, \textit{Negative}, \textit{Mixed\_feelings}, \textit{unknown\_state}, and \textit{not-Tamil}. 
Sentiment labels were assigned by the original corpus creators using human annotators; this work does not re-annotate sentiment. 
Our primary contribution to this dataset is a set of token-level language identification tags, generated by the superivsed fine-tuned model described in Section~4, which we apply to the romanized subset and which we make available as a resource for future research and downstream tasks.

Of the 44,161 utterances in the original dataset, 8,370 (19.0\%) contained Tamil Unicode characters. 
Since our language identification model is pre-trained on romanized (Latin-alphabet) text only, including Tamil-script data would drastically reduce the signal-to-noise ratio of the data and make training difficult. 
We therefore restrict analysis to the 35,650 (81.0\%) romanized utterances.

\subsection{Analysis Subset}

For the main statistical analysis, we further restrict to the three clearly delineated sentiment categories, excluding \textit{unknown\_state} (5,366) and \textit{not-Tamil} (2,079), yielding 28,205 utterances. 
Table~\ref{tab:dataset} reports descriptive statistics for this subset.

\begin{table}[!ht]
\centering
\small
\begin{tabular}{lrrrr}
\toprule
\textbf{Sentiment} & \textbf{n} & \textbf{EN\%} & \textbf{TA\%} & \textbf{Switches} \\
\midrule
Positive       & 19,871 & 34.3 & 61.7 & 1.99 \\
Mixed\_feelings & 3,855  & 28.5 & 67.4 & 2.50 \\
Negative       & 4,479  & 24.8 & 72.1 & 2.00 \\
\bottomrule
\end{tabular}
\caption{\label{tab:dataset}Dataset statistics by sentiment.}
\end{table}

Overall, Tamil is the matrix language (mean 62.8\% of tokens), with English as the embedded language (mean 30.2\%). 
Purely Tamil utterances account for the majority of the original dataset, confirming the matrix language assessment.

\section{Methods}

\subsection{Token-Level Language Identification}

To obtain per-utterance language measurements, we require reliable token-level language identification labels for the entire dataset. 
No existing annotated resource covers romanized English-Tamil code-mixed text, so we produced our own via supervised deep learning.
\footnote{Code and model parameters are available in \autoref{sec:code}}

One author manually annotated approximately 3,500 tokens (across $\sim$500 utterances) with one of three labels: 
\texttt{en} (English), \texttt{ta} (romanized Tamil), or \texttt{na} (numerals, ambiguous tokens, and punctuation). 
Ambiguous tokens constitute words like '\textit{mass}', which exists with the same spelling in both English and romanized Tamil, and could not always be readily disambiguated from sentence context during manual annotation. 
We include the option for \texttt{na}-labeled tokens to account for this ambiguity, providing the model with a way to acknowledge tokens of uncertain language identity; 
this results in greater confidence in the \texttt{en} and \texttt{ta} labels.
Annotation followed CoNLL format (one token per line, tab-separated label).

We fine-tuned \texttt{xlm-roberta-base} \citep{conneau2020unsupervised} for this token classification task. 
Training used a 80/20 train/validation split (approximately 1,789 / 221 training sentences) and the following hyperparameters: 
learning rate $2 \times 10^{-5}$, batch size 16, 5 epochs, linear warmup over 10\% of steps, weight decay 0.01. 
Only the first subword token receives a label for multi-subword words.
We experimented with including Tamil-script code-mixed text alongside the romanized data during fine-tuning; we found that validation-split performance decreased, likely due to the model's learned over-reliance on Tamil script as the sole indicator of Tamil language, preventing the model from learning to attend to morphosyntactic indicators of language identification.
Therefore, we consider only the romanized code-switching data in the following analysis.

After training, the model achieved an overall token-level accuracy of 93.3\% on the validation set, with a macro-averaged F1 of 0.844. 
Table~\ref{tab:langid_eval} reports per-class F1 scores. 
English and N/A tokens are classified with high accuracy; romanized Tamil is harder to distinguish from English at the token level, yielding a lower but still useful F1 of 0.638.

\begin{table}[!ht]
\centering
\small
\begin{tabular}{lc}
\toprule
\textbf{Class} & \textbf{F1} \\
\midrule
English (\texttt{en}) & 0.932 \\
Tamil (\texttt{ta})   & 0.638 \\
Ambiguous (\texttt{na}) & 0.963 \\
\midrule
Macro avg & 0.844 \\
Accuracy  & 0.933 \\
\bottomrule
\end{tabular}
\caption{\label{tab:langid_eval}Token-level language identification results on the validation set (epoch 5).}
\end{table}

The lower Tamil F1 (0.638) is expected due to our data filtering: transliterated Tamil words share the Latin script with English, so the model relies on learned spelling patterns and context rather than script identity to distinguish them. 
This ambiguity is present in our training data, since a single annotator without Tamil proficiency labeled tokens by lexical knowledge and limited document context, complicating borderline cases. 
Despite this, the model's high accuracy and English F1 indicate that the large majority of English tokens are identified correctly, which is the primary driver of our downstream proportion measurements.

We ran language-identification inference with our fine-tuned XLM-R model on all 35,650 Latin-script utterances in the subset, computing per-utterance English proportion, Tamil proportion, N/A proportion, and language switch count. 
Language switches are identified when consecutive tokens belong to different labels.

\subsection{Statistical Analysis}

We modeled two outcome variables using ordinary least-squares linear regression, with \textit{Positive} as the reference sentiment level (most frequent in the subset).
First, we explore whether sentiment is a significant predictor of English proportion in code-switching. 
We produced a linear model which predicts english proportion as a function of sentiment (Model 1a), and compared it against an extended model which accounts for interaction effects with sentence length (Model 1b).

Next, to investigate the effect of utterance sentiment on language switch frequency (rate of alternation between matrix and embedded language), we compare another set of linear models.
We modelled the number of switches per utterance as a function of sentiment (Model 2a), and compared against an interaction model which normalizes by sentence length (Model 2b).
Models were compared using ANOVA, and residual normality was confirmed via Q-Q plots.

\section{Results}

\subsection{RQ1: Sentiment and English Proportion}

Model 1a demonstrates that both \textit{Mixed\_feelings} ($\hat\beta = -0.057$, $p < 0.001$) and \textit{Negative} ($\hat\beta = -0.094$, $p < 0.001$) utterances have significantly lower English proportion than \textit{Positive} utterances, confirming a monotonic decrease in English use from positive to negative sentiment. 
However, the model's $R^2 = 0.01$ is low, indicating that sentiment alone explains only a small fraction of variance in English proportion, consistent with the high intra-group variability in the data (SD $\approx 24$--$26\%$ for all groups). 

Model 1b, which accounts for interaction with sentence length, improves $R^2$ to 0.02. 
However, the interaction terms are not significant ($p > 0.05$), indicating that utterance length does not moderate the sentiment effect on English proportion. 
An ANOVA comparison confirms that the interaction model provides a significantly better fit ($p < 0.05$), but the added complexity is modest. 
The effect of sentiment on English proportion is clear: positive utterances use 9.4 percent more English than negative utterances (34.3\% vs. 24.8\%).

\begin{figure}[!ht]
\centering
\includegraphics[width=\columnwidth]{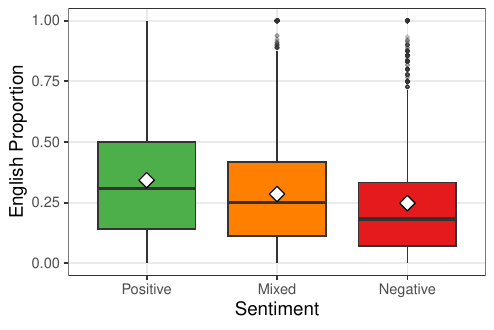}
\caption{Distribution of English proportion by sentiment class. White diamonds indicate group means.}
\label{fig:boxplot_en}
\end{figure}

\subsection{RQ2: Sentiment and Switch Frequency}

To quantify frequency of language alternation, we count a \textit{switch} as a whitespace-separated token boundary where the language ID labels of the left and right tokens are not the same, and neither label is \texttt{na}. 
This measures the number of times in an utterance when the speaker switches from English to Tamil or vice-versa, ignoring \texttt{na} tokens which are of ambiguous language, numerals, or punctuation.
In Model 2a, only \textit{Mixed\_feelings} shows a significant positive effect on switch frequency relative to \textit{Positive} ($\hat\beta = +0.51$, $p < 0.05$). 
However, the \textit{Negative} coefficient ($\hat\beta \approx 0.01$) is not significant ($p > 0.05$), so it should not be considered. 
Additionally, the $R^2 = 0.008$ is very low and plotted residuals deviate from normality, suggesting Model 2a is underspecified.

Adding the interaction with sentence length in Model 2b dramatically improves fit: $R^2$ increases from 0.008 to 0.30, and residual normality improves. 
All terms are significant ($p < 0.05$), including the interaction term for \textit{Mixed\_feelings}. 
This means the positive effect of mixed sentiment on switch frequency decreases slightly, but significantly, as utterance length increases. 
The \textit{Negative} coefficient becomes significant only in the interaction model, indicating that controlling for length is essential for detecting the negative sentiment effect on switch frequency.
These results confirm that mixed-sentiment utterances cause speakers to switch language more frequently than in entirely positive or negative ones, though the effect is slight.

\begin{figure}[!ht]
\centering
\includegraphics[width=\columnwidth]{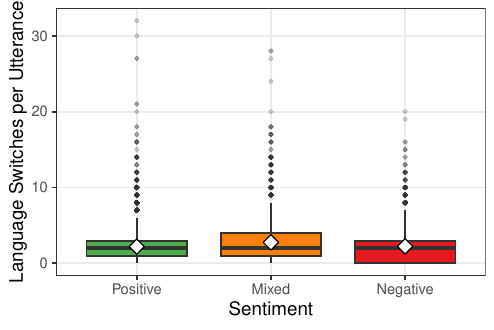}
\caption{Distribution of language switch count per utterance by sentiment class. White diamonds indicate group means.}
\label{fig:boxplot_switches}
\end{figure}

\section{Discussion and Conclusion}

Our findings provide quantitative support for the sociolinguistic hypothesis that emotional content influences language choice in English-Tamil code-switched speech. 
Positive utterances show significantly greater English proportion, consistent with English carrying academic and professional prestige as a colonial legacy in southern India. 
Negative utterances show greater Tamil use, consistent with Tamil serving as the language of emotional expression and intimacy for native speakers.
Moreover, the findings support our intuition that, if English and Tamil are associated with opposite sentiments, then mixed-sentiment code-switching utterances demonstrate the highest frequency of language switches.

The finding that mixed-sentiment utterances hold the highest switching rate aligns with the characterization of intra-discursive code-mixing as an unconscious process tied to internal states \citep{muysken2000bilingual}. 
When a speaker holds ambivalent or conflicting feelings, navigation between emotional registers may trigger more frequent, unplanned language alternation than a uniformly negative affect, which could instead produce sustained Tamil use without switching.

Despite these clear findings, the low $R^2$ values for both proportion models ($\sim$0.01--0.02) indicate the sociolinguistic complexity of code-switching: 
sentiment is one factor among many (domain, register, audience, platform) that influence a speaker's language of expression, and more work remains to fully understand the often subconscious practice of code-switching. 

\section{Limitations and Future Work}

Several limitations apply: 
First, restricting analysis to romanized text excludes 19\% of the corpus, so Tamil-script utterances may exhibit different patterns. 
Second, there is substantial class imbalance (Positive is $\sim$70\% of the analysis set), which may inflate the apparent significance of the reference-level effect. 
Third, the manual language ID training set we produced is small ($\sim$3,500 tokens), and annotation was performed by a single annotator with imperfect grasp of Tamil. 

Future work should extend this analysis to other prestige-heritage language pairs with comparable sociolinguistic dynamics (e.g., French-Arabic, Spanish-Tagalog), and apply character-level or script-aware models to handle Tamil-script data.



\section{Ethics Statement}

The YouTube comment data used in this work was collected and annotated by \citet{chakravarthi2022dravidian} and is publicly available. 
We use only the pre-existing text and sentiment annotations. 
No new user data was collected, and no individuals are identified or targeted in our analysis. 
We acknowledge that social media text may contain offensive or sensitive content, but this work analyzes data at the corpus level and does not present or reproduce individual harmful utterances. 
Manual annotation was performed by a single annotator and reflects one individual's linguistic knowledge and implicit bias. 
Downstream analyses should account for the resulting uncertainty, particularly for Tamil token classification.

\section{Data and Code Availability}\label{sec:code}

The full DravidianCodeMix dataset used in this work is available via the original authors' publication \citep{chakravarthi2022dravidian}. 
Our fine-tuned XLM-RoBERTa language identification model, its associated training code, our manually-annotated training data, and the resulting token-level annotations for the romanized subset are available at the following GitHub repository: \href{https://github.com/paulbontempo/en-ta-cs.git}{en-ta-cs repo}.

\section{References}\label{sec:reference}

\bibliographystyle{lrec2026-natbib}
\bibliography{entacs}

@article{chakravarthi2022dravidian,
  author    = {Chakravarthi, Bharathi Raja and
               Priyadharshini, Ruba and
               Muralidaran, Vigneshwaran and
               Jose, Navya and
               Suryawanshi, Shardul and
               Sherly, Elizabeth and
               McCrae, John P.},
  title     = {{DravidianCodeMix}: Sentiment Analysis and Offensive Language Identification Dataset for {Dravidian} Languages in Code-Mixed Text},
  journal   = {Language Resources and Evaluation},
  year      = {2022},
  month     = {Feb},
  doi       = {10.1007/s10579-022-09583-7},
  url       = {https://doi.org/10.1007/s10579-022-09583-7}
}

@inproceedings{conneau2020unsupervised,
  author    = {Conneau, Alexis and
               Khandelwal, Kartikay and
               Goyal, Naman and
               Chaudhary, Vishrav and
               Wenzek, Guillaume and
               Guzm{\'{a}}n, Francisco and
               Grave, Edouard and
               Ott, Myle and
               Zettlemoyer, Luke and
               Stoyanov, Veselin},
  title     = {Unsupervised Cross-lingual Representation Learning at Scale},
  booktitle = {Proceedings of the 58th Annual Meeting of the Association for Computational Linguistics},
  year      = {2020},
  pages     = {8440--8451},
  publisher = {Association for Computational Linguistics},
  doi       = {10.18653/v1/2020.acl-main.747}
}

@article{poplack1980sometimes,
  author    = {Poplack, Shana},
  title     = {Sometimes {I'll} Start a Sentence in {Spanish} {Y} Termino en {Espa\~{n}ol}: Toward a Typology of Code-Switching},
  journal   = {Linguistics},
  volume    = {18},
  pages     = {581--618},
  year      = {1980},
  publisher = {De Gruyter Mouton}
}

@book{muysken2000bilingual,
  author    = {Muysken, Pieter},
  title     = {Bilingual Speech: A Typology of Code-Mixing},
  publisher = {Cambridge University Press},
  year      = {2000},
  address   = {Cambridge}
}

@article{Sarma2025language,
  title = {Intersections between heritage, multilingualism, and education: language acquisition in India},
  volume = {19},
  ISSN = {1662-5161},
  url = {http://dx.doi.org/10.3389/fnhum.2025.1538482},
  DOI = {10.3389/fnhum.2025.1538482},
  journal = {Frontiers in Human Neuroscience},
  publisher = {Frontiers Media SA},
  author = {Sarma,  Vaijayanthi M.},
  year = {2025},
  month = oct 
}

@article{Sankoff1981,
  title = {A formal grammar for code‐switching},
  volume = {14},
  ISSN = {0031-1251},
  url = {http://dx.doi.org/10.1080/08351818109370523},
  DOI = {10.1080/08351818109370523},
  number = {1},
  journal = {Paper in Linguistics},
  publisher = {Informa UK Limited},
  author = {Sankoff,  David and Poplack,  Shana},
  year = {1981},
  month = jan,
  pages = {3–45}
}

@article{Eldho2023english,
  title = {Choice of language in the construction of cultural identity by Tamil speakers in India},
  volume = {10},
  ISSN = {2214-3165},
  url = {http://dx.doi.org/10.1075/ijolc.00045.eld},
  DOI = {10.1075/ijolc.00045.eld},
  number = {1},
  journal = {International Journal of Language and Culture},
  publisher = {John Benjamins Publishing Company},
  author = {Eldho,  Elizabeth and Kumar,  Rajesh},
  year = {2023},
  month = dec,
  pages = {54–86}
}

@inproceedings{babu2021sentiment,
  author    = {Babu, Yandrapati Prakash and Rajagopal, Eswari},
  title     = {Sentiment Analysis on {Dravidian} Code-Mixed {YouTube} Comments using Paraphrase {XLM-RoBERTa} Model},
  booktitle = {Proceedings of the Forum for Information Retrieval Evaluation (FIRE)},
  year      = {2021},
  url       = {https://api.semanticscholar.org/CorpusID:251322714}
}

@inproceedings{sitaram2019survey,
  author    = {Sitaram, Sunayana and
               Chandu, Khyathi Raghavi and
               Rallabandi, SaiKrishna and
               Black, Alan W.},
  title     = {A Survey of Code-switched Speech and Language Processing},
  booktitle = {Proceedings of the 57th Annual Meeting of the Association for Computational Linguistics},
  year      = {2019},
  note      = {arXiv:1904.00784}
}

@inproceedings{aguilar2020lince,
  author    = {Aguilar, Gustavo and
               AlGhamdi, Fahad and
               Soto, Victor and
               Solorio, Thamar and
               Diab, Mona and
               Ghazaleh, Vasileios},
  title     = {{LinCE}: A Centralized Benchmark for Linguistic Code-switching Evaluation},
  booktitle = {Proceedings of the 12th Language Resources and Evaluation Conference},
  year      = {2020},
  pages     = {1803--1812},
  publisher = {European Language Resources Association}
}

@inproceedings{yoder2017code,
  author    = {Yoder, Michael and
               Rijhwani, Shruti and
               Ros{\'e}, Carolyn and
               Levin, Lori},
  title     = {Code-Switching as a Social Act: The Case of {Arabic} {Wikipedia} Talk Pages},
  booktitle = {Proceedings of the Second Workshop on Natural Language Processing and Computational Social Science},
  year      = {2017},
  pages     = {73--82}
}

@book{MyersScotton2002,
  title = {Contact Linguistics},
  ISBN = {9780198299530},
  url = {http://dx.doi.org/10.1093/acprof:oso/9780198299530.001.0001},
  DOI = {10.1093/acprof:oso/9780198299530.001.0001},
  publisher = {Oxford University Press},
  author = {Myers-Scotton,  Carol},
  year = {2002},
  month = aug 
}

\end{document}